\documentclass{article}

\usepackage[final]{neurips_2025}

\usepackage[utf8]{inputenc} 
\usepackage[T1]{fontenc}    
\usepackage{hyperref}       
\usepackage{url}            
\usepackage{booktabs}       
\usepackage{amsfonts}       
\usepackage{nicefrac}       
\usepackage{microtype}      
\usepackage{xcolor}         
\usepackage{graphicx}
\usepackage{enumitem}
\usepackage{multirow}
\title{AgriDoctor: A Multimodal Intelligent \\Assistant for Agriculture}

\author{
  Mingqing Zhang\textsuperscript{1,2}\thanks{$\ $  The first two authors contributed equally to this work.},
  Zhuoning Xu\textsuperscript{1,2}\footnotemark[1],
  Peijie Wang\textsuperscript{1,2},
  Rongji Li\textsuperscript{1,2},
  Liang Wang\textsuperscript{1,2}, \AND
  Qiang Liu\textsuperscript{1,2}, 
  Jian Xu\textsuperscript{1,2},
  Xuyao Zhang\textsuperscript{1,2},
  Shu Wu\textsuperscript{1,2},
  Liang Wang\textsuperscript{1,2} \\
  \textsuperscript{1}
  State Key Laboratory of Multimodal Artificial Intelligence Systems \\
  Institute of Automation, Chinese Academy of Sciences \\
  \textsuperscript{2}School of Artificial Intelligence, University of Chinese Academy of Sciences \\
  \texttt{\{mingqing.zhang, liang.wang\}@cripac.ia.ac.cn}, \\
  \texttt{\{xuzhuoning2023, wangpeijie2023, lirongji2025, jian.xu\}@ia.ac.cn}, \\
  \texttt{\{qiang.liu, xyz, shu.wu, wangliang\}@nlpr.ia.ac.cn}
}

\begin{document}

\maketitle

\begin{abstract}
Accurate crop disease diagnosis is essential for sustainable agriculture and global food security. Existing methods, which primarily rely on unimodal models such as image-based classifiers and object detectors, are limited in their ability to incorporate domain-specific agricultural knowledge and lack support for interactive, language-based understanding.
Recent advances in large language models (LLMs) and large vision-language models (LVLMs) have opened new avenues for multimodal reasoning. However, their performance in agricultural contexts remains limited due to the absence of specialized datasets and insufficient domain adaptation.
In this work, we propose \textbf{AgriDoctor}, a modular and extensible multimodal framework designed for intelligent crop disease diagnosis and agricultural knowledge interaction. As a pioneering effort to introduce agent-based multimodal reasoning into the agricultural domain, AgriDoctor offers a novel paradigm for building interactive and domain-adaptive crop health solutions. It integrates five core components: a router, classifier, detector, knowledge retriever and LLMs. 
To facilitate effective training and evaluation, we construct \textbf{AgriMM}, a comprehensive benchmark comprising 400{,}000 annotated disease images, 831 expert-curated knowledge entries, and 300{,}000 bilingual prompts for intent-driven tool selection.
Extensive experiments demonstrate that AgriDoctor, trained on AgriMM, significantly outperforms state-of-the-art LVLMs on fine-grained agricultural tasks, establishing a new paradigm for intelligent and sustainable farming applications.

\end{abstract}

\section{Introduction}

Accurate identification of crop diseases plays a pivotal role in promoting sustainable agriculture, ensuring crop productivity, and maintaining global food security. However, most existing crop disease recognition approaches are based on unimodal techniques, such as image-based classification models \cite{agarwal2020potato,khamparia2020seasonal,sakkarvarthi2022detection,dai2024dfn} and object detection frameworks \cite{saleem2019plant,panchal2023image,wang2024pd}.While effective in visual recognition, these methods lack deep modeling of agricultural knowledge and cannot interactively understand natural language queries, limiting their real-world use in agriculture.

In recent years, artificial intelligence has undergone rapid development, particularly in the field of natural language processing (NLP). Among these advancements, the emergence of LLMs has further accelerated progress. Trained on massive textual corpora, LLMs have demonstrated remarkable capabilities in language understanding and generation tasks \cite{y2022large,gandhi2023understanding,nam2024using,du2024haloscope,liu2025llm4gen}. Under instruction-based paradigms, LLMs can generate informative, contextually relevant responses grounded in domain knowledge, enabling natural language interaction with users. They have been successfully applied to various real-world domains, including legal consultation \cite{cheong2024not,guha2023legalbench,bignotti2024legal} and clinical decision support \cite{li2024mediq,kweon2024ehrnoteqa,singhal2023large}. In the field of crop science, preliminary efforts have explored the use of LLMs for tasks such as answering pest and disease-related questions or solving agricultural examination problems \cite{zhang2024empowering,shaikh2024role,kuska2024ai}. Despite these developments, current LLMs still exhibit significant limitations in handling agricultural-specific terminology, multilingual capabilities, and external knowledge integration. Their performance in multimodal agricultural scenarios involving both images and text remains suboptimal.

To bridge this gap, the emergence of LVLMs has brought significant advances in the field of multimodal learning. Models such as GPT-4o~\cite{hurst2024gpt}, Flamingo~\cite{alayrac2022flamingo}, Llava-onevision~\cite{li2024llava}, and Qwen2.5-VL~\cite{bai2025qwen2} adopt unified architectures that integrate visual and textual processing, achieving impressive results in open-domain tasks such as image captioning, visual question answering, and multi-turn dialogue~\cite{deng2024enhancing,hu2024matryoshka,liu2024convbench}. However, their applications in the agricultural domain remain limited. These models are predominantly trained on general-purpose datasets and lack effective modeling of agriculture-specific visual concepts and expert knowledge. As a result, they face significant challenges in fine-grained disease classification, precise lesion localization, and domain-specific question answering within agricultural contexts.

In this work, we introduce \textbf{AgriDoctor}, a domain-adapted, modular, and extensible multimodal framework specifically designed for intelligent crop disease diagnosis and agricultural knowledge interaction. As a pioneering attempt to apply agent-based multimodal reasoning in the agricultural domain, AgriDoctor integrates a unified pipeline comprising five core components: a router for intent classification and task dispatching, a disease classifier, a lesion detector, a knowledge retriever, and a LLM that integrates upstream visual outputs with textual inputs for context-aware reasoning and response generation. The model supports both pure textual queries and multimodal inputs, and produces multimodal outputs, including classification results, lesion visualizations, and textual explanations.

Crucial to the effectiveness of AgriDoctor is the \textbf{AgriMM} dataset, a large-scale, high-quality multimodal agricultural benchmark we construct to support fine-grained reasoning. AgriMM includes 400{,}000 expert-annotated disease images across 29 crops and 138 disease types, bounding-box annotations for lesion localization, a structured knowledge base covering 831 pest and disease entries in both English and Chinese, and 300{,}000 bilingual prompt examples for intent-oriented tool selection. This rich and well-structured dataset enables AgriDoctor to learn fine-grained visual patterns, domain-specific knowledge, and task-aware reasoning strategies essential for real-world agricultural applications.

Through comprehensive experiments, we demonstrate that AgriDoctor, when trained with AgriMM, significantly outperforms general-purpose LVLMs on a variety of agricultural tasks, including disease classification, lesion detection, and knowledge-grounded question answering. Our results highlight the importance of domain adaptation, modular reasoning, and high-quality agricultural supervision for advancing multimodal AI in agriculture. We hope that AgriDoctor and AgriMM together will serve as strong foundations for future research on domain-specific LVLMs and practical AI applications in sustainable farming.

The key contributions of this work are summarized as follows:
\begin{itemize}[leftmargin=*]
    \item We propose AgriDoctor, a pioneering agent-style multimodal framework that combines image understanding and natural language reasoning through a modular and extensible design, enabling intelligent agricultural assistance across diverse input modalities.
    
    \item We construct AgriMM, a comprehensive and large-scale benchmark for multimodal agricultural tasks, comprising 400{,}000 annotated disease images, 831 expert-curated entries of agricultural pest and disease knowledge, and 300{,}000 prompts for intent-driven tool selection.
    
    \item We perform extensive experiments demonstrating that AgriDoctor, trained on AgriMM, outperforms state-of-the-art LVLMs on fine-grained agricultural tasks, setting a new standard for intelligent farming applications.
\end{itemize}

\section{Related Work}
\subsection{Tool Use in Multimodal Agents}
Tool learning has become essential for enhancing agents’ abilities to handle complex multimodal tasks. Current approaches fall into two main paradigms: retriever-based and LLM-based tool selection~\cite{qu2024tool}. Retriever-based methods match tools via rules or vector retrieval, while LLM-based approaches dynamically select tools based on semantic understanding.
Recent work shows tool learning's promise in multimodal contexts. For example, LLAVA-PLUS~\cite{liu2025llava} parses user instructions to orchestrate tool use, and CLOVA~\cite{gao2024clova} introduces a three-stage framework (inference, reflection, learning) to support continuous tool adaptation. To evaluate such capabilities, benchmarks like T-Eval~\cite{chen2023t}, API-Bank~\cite{li2023api}, and CARP~\cite{zhang2024evaluating} provide standardized criteria across dimensions such as reasoning, planning, and API interaction.
However, relying on LLMs as tool selectors introduces challenges: 1) hallucinations can cause incorrect or conflicting tool use~\cite{wang2023mint, gou2023critic, liu2025llava, gao2024clova, 2023toolformer, 2023toolkengpt}; 2) performance heavily depends on model quality and prompt design; and 3) the inference cost of large models (e.g., Qwen2.5-7B takes ~420ms per call) limits applicability in real-time domains like agriculture. To mitigate these issues, we propose AgriDoctor.

\subsection{Agricultural LLMs}

In recent years, LLMs have been increasingly applied to agricultural knowledge services, incorporating expert feedback and domain-specific tuning to address agricultural challenges~\cite{silva2023gpt}. Techniques such as Retrieval-Augmented Generation and Ensemble Refinement have further improved generative performance. Instruction-tuned LLaMA 2 models, trained on over 1.5 million plant science papers, enhance understanding of plant science topics~\cite{yang2024pllama, llama2}.
Multimodal models using Vision-Language Pretraining have achieved high accuracy (e.g., 94.84\% on cucumber disease recognition) by learning semantic correlations across modalities~\cite{cao2023cucumber}. These efforts, combining domain-specific pretraining and knowledge augmentation, significantly advance LLMs' capabilities in agricultural reasoning, multimodal fusion, and information extraction~\cite{agricultural2023llm}.
Nonetheless, current models largely focus on single-task solutions. A unified architecture and standard benchmarks for complex agricultural tasks are still lacking. To address this, we propose AgriDoctor, A multimodal intelligent
assistant for agriculture.
\section{AgriMM}

Despite the remarkable performance of contemporary vision-language models and general-purpose multimodal large models on public benchmarks, they often exhibit limitations in fine-grained reasoning and domain-specific applications, such as agricultural disease diagnosis and plant pathology detection. These limitations are primarily attributed to the scarcity of high-quality, domain-annotated datasets that capture the complexity and variability inherent in real-world agricultural scenarios.
In this work, we construct a large-scale multimodal agricultural dataset, \textbf{AgriMM}, which contains fine-grained annotations for 400,000 disease-related images, structured expert knowledge covering 831 types of crop pests and diseases, and 300,000 intent-oriented question-answer pairs focused on plant pathology. 
In this section, we describe the construction of AgriMM, specifically developed to address this gap by providing fine-grained annotations of crop leaf diseases, encompassing both classification labels and localization information. The images were systematically curated and annotated under the supervision of agricultural experts to ensure label accuracy and precise spatial alignment of disease symptoms. This dataset is designed to facilitate multimodal learning tasks, including disease classification, lesion localization, and knowledge-grounded question answering within the agricultural domain.
\subsection{Image Data}

The image dataset utilized in this study comprises two primary sources: 

(1) web data, including agricultural disease images obtained from public platforms such as Kaggle and Roboflow~\cite{classification-of-disease-dataset, wheat-crop-disease-detection-dataset, wheatleafdisease-dataset, apple-treed-dataset, apple-disease-detection2-dataset, apple-diseases-detection-1po19-dataset, fp-dgnze-dataset, tomato-disease-bj8e6-dataset, tomato-clean-dataset-dataset, plant-diagnosis-5--mango-and-family-dataset, mango-diseases-dataset, coffee-reazm-dataset, tea-disease-zfekj-dataset, tea-diseases-db19c-dataset}, as well as large-scale crop image collections gathered via web crawling techniques;

(2) private data, consisting of plant disease images captured through field surveys conducted in real-world agricultural environments.

All images were annotated with assistance from agricultural experts, who provided accurate labels for both crop categories and disease types.  
Through large-scale data collection and expert-guided annotation, a comprehensive dataset of approximately 400{,}000 images was constructed, covering 29 distinct crop categories and 138 types of agricultural diseases.

\begin{figure*}[t]
\setlength{\abovecaptionskip}{0pt}
\setlength{\belowcaptionskip}{0pt}
\begin{center}
\includegraphics[width=1\textheight, angle=-90]{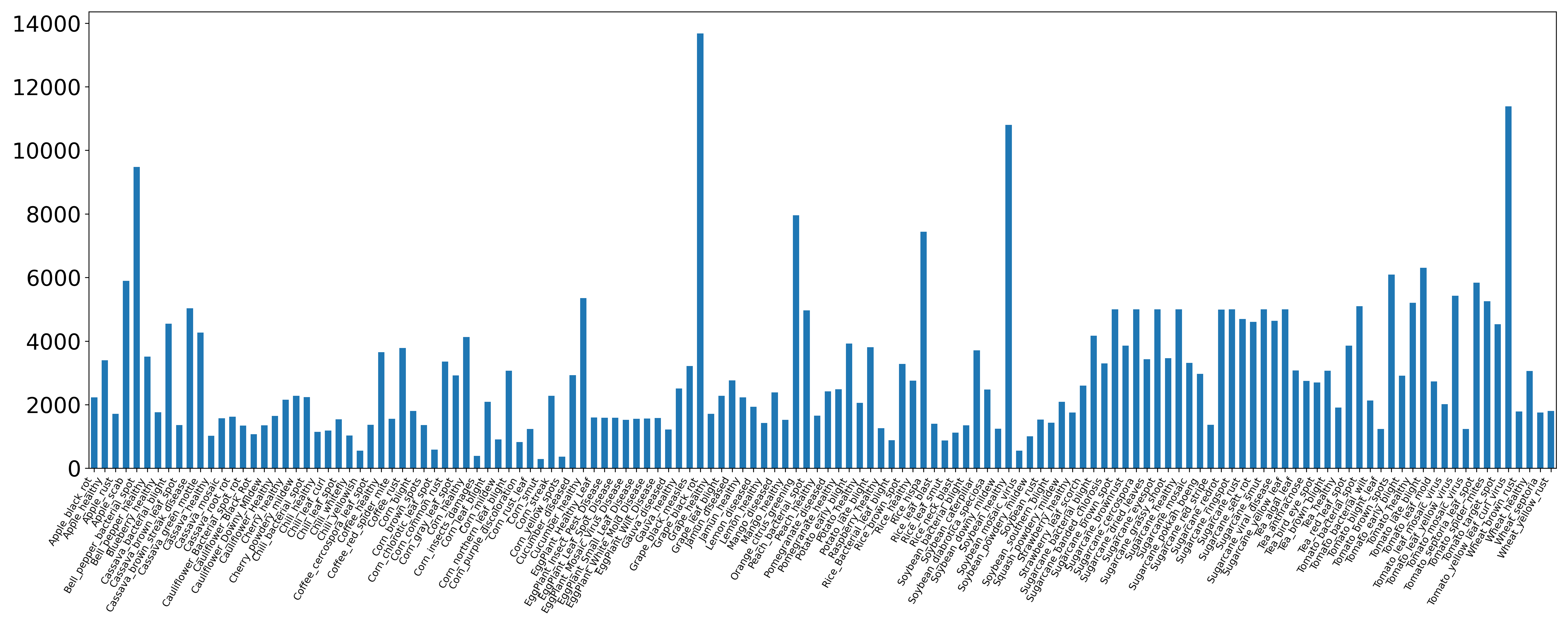}
\end{center}
\vspace{0cm}
\caption{Distribution of the Images}
\label{fig:2}
\end{figure*}

In our dataset, 128 disease categories comprise more than 1{,}000 images each, whereas the remaining 10 categories contain between 200 and 1{,}000 images, as illustrated in Figure~\ref{fig:2}.
Furthermore, to facilitate object detection tasks, we manually annotated the disease-infected regions across a substantial portion of the dataset.  
All bounding boxes adhere to the standard YOLO annotation format (\texttt{<class\_id> <x\_center> <y\_center> <width> <height>}) with coordinates normalized to the image dimensions. 

\subsection{Disease and Pest Expertise}
To enhance the model’s capability in reasoning and responding to agricultural health-related tasks, we constructed a high-quality knowledge base encompassing both crop diseases and pests. This knowledge base comprises a total of 831 meticulously curated entries, each providing detailed bilingual (Chinese and English) textual descriptions. The content covers expert-level information, including disease symptoms, transmission mechanisms, physiological impacts, and integrated management strategies.

The knowledge base not only comprehensively covers all disease categories represented in the Agricultural Disease Image Dataset, but also extends to include a substantial amount of pest and disease information not directly captured in the image data. This significantly enhances the model’s generalization and response ability in non-visual input scenarios.
All textual entries are derived from peer-reviewed agricultural research publications, authoritative expert manuals, and curated agricultural knowledge databases. Each entry has been systematically categorized and standardized to facilitate seamless integration into downstream retrieval modules. This structured knowledge serves as essential external semantic context, enabling the model to perform tasks such as pest and disease control recommendation and general agricultural question answering.

\subsection{Tool Selection Data}

To effectively support downstream task processing, we constructed a bilingual prompt dataset in both Chinese and English. The tool selection framework is organized around three distinct task categories, each specifically designed to address diverse agricultural query types.

To ensure comprehensive coverage of real-world scenarios, we employed GPT-4o to generate a large corpus of candidate prompts. Following a rigorous process of manual filtering and verification, we curated a high-quality dataset consisting of approximately 300{,}000 prompt instances. Each instance is annotated with a corresponding task label, enabling supervised learning and accurate task alignment.

\textbf{Agricultural Knowledge Retrieval}  
This category is intended for queries that require domain-specific knowledge interpretation. It includes \textit{disease-specific inquiries}, \textit{general agricultural knowledge questions}, and \textit{expert-level information requests unrelated to classification or detection}.

\textbf{Disease Classification}  
This category focuses on the categorization of crop diseases. It covers \textit{pure classification tasks}, \textit{disease identification requests}, and \textit{symptom-based classification queries}.

\textbf{Disease Detection}  
This category targets detection-oriented queries, including \textit{pure detection tasks}, \textit{combined classification and detection tasks}, and \textit{queries involving localization and identification of affected regions}.

The annotated dataset facilitates fine-grained understanding of user intent and supports dynamic routing to appropriate task-specific modules such as \textit{knowledge retrieval}, \textit{disease classification}, or \textit{region-level detection}. This design ensures precise handling of multimodal agricultural inputs and promotes effective integration of expert knowledge into the model’s reasoning process.


\section{AgriDoctor}

\begin{figure*}[t]
\setlength{\abovecaptionskip}{0pt}  
\setlength{\belowcaptionskip}{0pt}  
   \begin{center}
\includegraphics[width=1\textwidth]{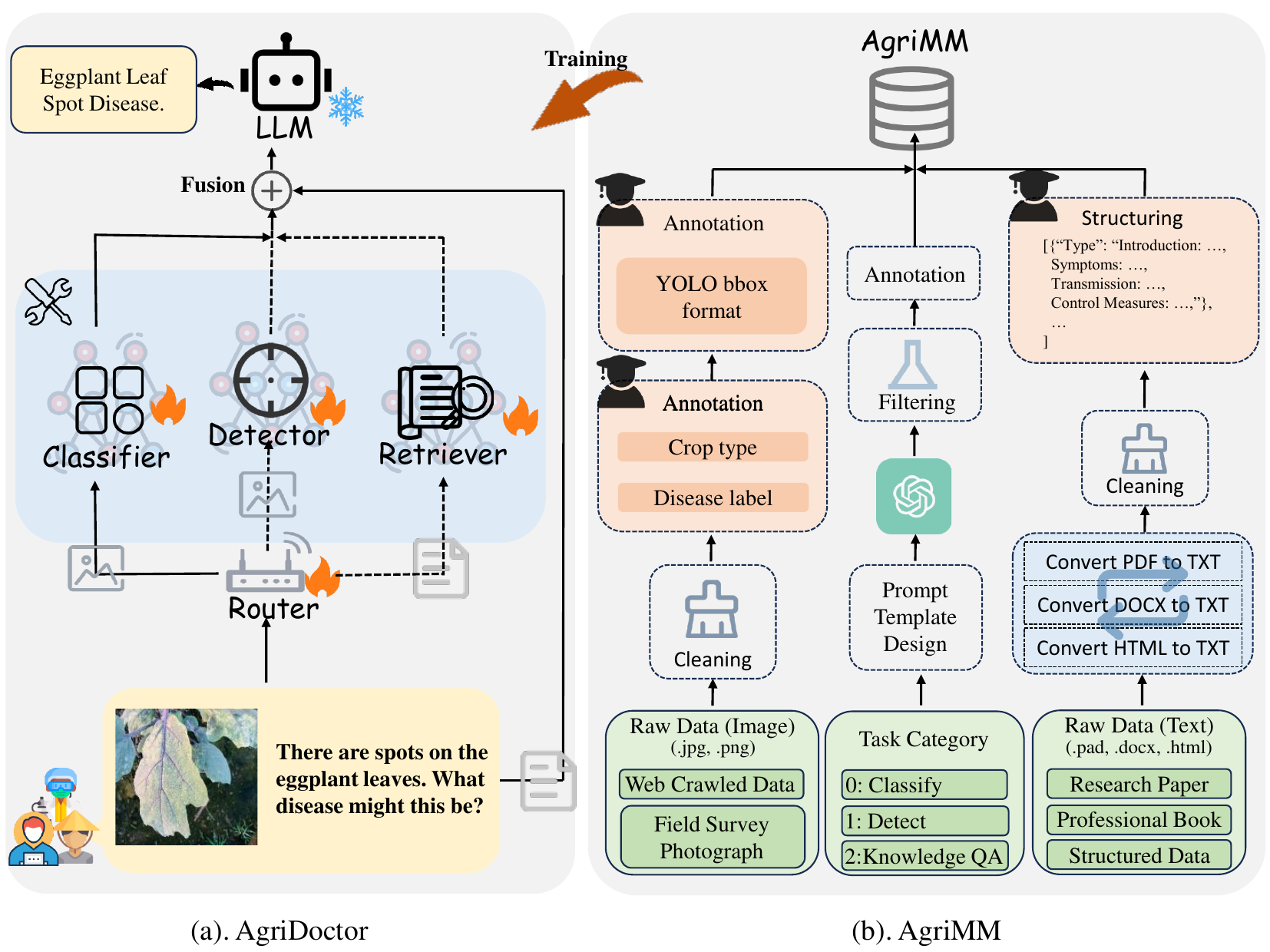}
   \end{center}
   \caption{(a) The overall architecture of AgriDoctor, a modular multimodal agent that routes user queries through disease classification, lesion detection, or knowledge retrieval tools, followed by LLM-based output fusion.(b) The data construction workflow of AgriMM, including image annotation and knowledge structuring from raw visual and textual sources. This dataset supports the training of AgriDoctor's downstream modules.
} 
   \label{fig:1}
\end{figure*}

We propose AgriDoctor, a comprehensive framework comprising five core components: a router for tool selection, a disease classifier, a disease detector, a knowledge retriever, and a LLM. This architecture adopts a modular and extensible pipeline design, enabling accurate disease analysis and expert-level knowledge dissemination, as illustrated in Figure \ref{fig:1}.

\subsection{Router}
Within the AgriDoctor framework, the Router module serves as a central decision-making component, with the primary objective of accurately identifying user intent and routing multimodal inputs to the most appropriate downstream task modules. This enables intelligent parsing and resolution of diverse agricultural queries in real-world settings.
When processing multimodal inputs that contain both textual and visual information, the Router module performs three essential operations: language detection, intent classification, and task routing. The workflow is detailed as follows:

\textit{Language Detection and Model Selection:}  
To enhance intent recognition performance across multilingual scenarios (e.g., Chinese and English), we constructed and fine-tuned separate BERT-based classification models using language-specific corpora. The system first detects the language of the user's textual input via a lightweight language identifier, and subsequently selects the corresponding BERT model for downstream semantic understanding and intent inference.

\textit{Intent Recognition and Task Classification:}  
Based on the output of the language-specific BERT classifier, the Router module embeds the input text and predicts one of several predefined intent categories. The system currently supports the following three primary tasks:
\begin{itemize}[leftmargin=*]
    \item \textit{Disease Classification:} For queries that aim to identify the type of disease present in an image (e.g., "What disease is this?"), the system activates the disease classification module to perform fine-grained visual categorization.
    \item \textit{Disease Detection:} For requests focused on localizing symptomatic regions within an image (e.g., "Please highlight the diseased area"), the system invokes the object detection module to predict bounding boxes around lesion-affected areas.
    \item \textit{Knowledge-Grounded Question Answering:} For questions centered on non-visual agricultural knowledge such as disease transmission mechanisms, physiological impacts, or control strategies, the system routes the input to the knowledge retrieval module, which combines user intent with curated domain knowledge to generate expert-level responses.
\end{itemize}

Once intent classification is completed, the Router automatically dispatches the input to the corresponding task-specific tool for processing. In summary, by integrating language-aware intent recognition with a modular task routing strategy, the Router significantly enhances the system's ability to handle multilingual, multimodal, and multitask inputs, making it a key enabler of AgriDoctor’s intelligent agricultural diagnosis and question answering capabilities.

\subsection{Disease Classification Tool}
The disease classification tool in the AgriDoctor framework is responsible for identifying the specific disease type present in a given crop image. We adopt the CLIP~\cite{radford2021learning} model with a Vision Transformer (ViT)~\cite{dosovitskiy2020image} backbone as a fixed visual encoder to extract high-level image features. Instead of using prompt-based matching, we append a trainable classification head on top of the vision encoder and fine-tune it in a supervised manner using labeled disease images. This approach enables robust and scalable classification across diverse crop diseases without requiring handcrafted features or bounding box annotations.

\subsection{Disease Detection Tool}
The disease detection tool in the AgriDoctor system is designed to achieve high-precision spatial localization of lesion regions within crop images. To accomplish this, we adopt YOLOv12 as the core detection model, which is a state-of-the-art one-stage object detection architecture capable of identifying and delineating disease-affected areas under complex agricultural conditions.

YOLOv12~\cite{tian2025yolov12} incorporates multi-scale feature aggregation, transformer-based attention enhancement mechanisms, and adaptive label assignment strategies, which significantly improve detection accuracy. These enhancements are especially effective in cases where lesions are small, occluded, or visually subtle.
The tool is trained on our carefully curated AgriMM dataset. During inference, the model takes a crop image as input and outputs a set of bounding boxes, each associated with a confidence score and a corresponding disease class label, enabling automatic localization and classification of lesion regions.

\subsection{Knowledge Retrieval Tool}
The knowledge retrieval tool in AgriDoctor is designed to retrieve domain-specific agricultural knowledge relevant to a user's textual input. It functions as one of the core tools within the AgriDoctor framework and serves as the entry point for handling non-visual, knowledge-intensive queries. Rather than performing reasoning directly, this module focuses on identifying and returning textual content that semantically aligns with the input query.

The underlying knowledge base is constructed from our AgriMM dataset and consists of 831 carefully curated bilingual (Chinese and English) knowledge entries. Each entry captures expert-level agricultural information, including crop disease characteristics, pest types, symptom descriptions, transmission pathways, and integrated control measures. All textual content is stored in a structured format, allowing for efficient semantic retrieval across diverse agricultural scenarios.

To perform retrieval, we first apply keyword extraction techniques—using the Jieba toolkit—for lightweight query summarization. The processed query is then embedded into a dense vector representation using a pre-trained BERT model, selected based on the input language. Each knowledge entry in the database is similarly encoded using mean pooling over token embeddings. The entire set of vectors is indexed using FAISS with an $L_2$ similarity metric, enabling fast and accurate retrieval over high-dimensional semantic spaces.

Upon receiving a query, the system retrieves the top-$k$ knowledge entries that are most relevant based on embedding similarity. These entries are not immediately used to generate a final answer; instead, they serve as the semantic context for downstream reasoning. 
By decoupling knowledge retrieval from language generation, this tool provides a scalable and modular mechanism to ground AgriDoctor’s responses in high-quality, domain-specific information, while maintaining extensibility across both languages and agricultural subdomains.

\subsection{Output Fusion}
To enable coherent reasoning and generate contextually appropriate responses, AgriDoctor integrates the outputs of its modular tools with the user query through a unified output fusion process. This fusion step is performed by the LLM, which serves as the central reasoning engine responsible for interpreting the tool outputs in relation to the user’s intent.

At each interaction step, only one tool is invoked based on the result of intent classification. The output of the selected tool, whether it is a disease label, a set of bounding boxes, or a retrieved knowledge snippet, is then integrated into a structured prompt that combines the original user input with the corresponding tool-specific result.


This approach allows the language model to synthesize multimodal information or domain-specific content without requiring changes to its internal architecture. All fusion is conducted at the prompt level, ensuring interpretability and preserving modularity across the system.

By leveraging LLM for this output fusion task, AgriDoctor achieves flexible and extensible reasoning capabilities across heterogeneous tool outputs. The fusion process not only ensures that responses are consistent with intermediate results, but also allows seamless scaling to more complex workflows involving new tools or modalities in future versions of the system.

\section{Experiments}
\subsection{Tool training}
To enable AgriDoctor to achieve optimal performance, we first conducted dedicated training of each functional module using the AgriMM dataset. The dataset was split into training, validation, and test sets following an 8:1:1 ratio.

\textbf{Router} We trained and evaluated separate BERT models on the Chinese and English intent classification subsets of the AgriMM dataset. The experimental results demonstrate that BERT achieves an accuracy of 99.00\% on English tasks and 98.82\% on Chinese tasks. These results indicate that our fine-tuned BERT models exhibit strong performance across both linguistic settings. In addition to its high accuracy, BERT’s significantly smaller model size enhances its practicality for tool selection tasks. Specifically, the lightweight architecture of BERT enables substantial reductions in both training and inference costs while maintaining robust performance. By integrating BERT into the AgriDoctor framework, we achieve highly accurate intent understanding, which in turn facilitates effective coordination and optimization among the system’s modular tools.

\textbf{Disease Classification Tool}
This tool is designed to train the crop disease classification tool. We adopt a transfer learning strategy by freezing the visual encoder of the CLIP model and fine-tuning only the classification head to adapt to the specific task. The model is trained on a single A6000 GPU using the Adam optimizer with a learning rate of 0.0001 and a batch size of 128, targeting 138 disease categories. Training converges around the 15th epoch, and the model weights are saved at the point of best validation performance. Experimental results demonstrate that the pretrained visual features can be effectively transferred to the agricultural disease recognition task, achieving a classification accuracy of 96.2\% on the test set, indicating strong generalization ability and robust recognition performance.

\textbf{Disease Detection Tool}
This tool is dedicated to the task of crop disease object detection. We adopt the YOLOv12 model and train it on four A6000 GPUs for 200 epochs, with a batch size of 128 and an initial learning rate of 0.02. The total training time was approximately 280 hours. During evaluation, we employed standard object detection metrics, including Precision, Recall, mAP@50 (mean Average Precision at an IoU threshold of 0.5), and mAP@50–95 (mean Average Precision across multiple IoU thresholds from 0.5 to 0.95).
Precision measures the proportion of correctly identified objects among all predictions, while Recall reflects the model’s ability to detect all ground-truth objects. mAP serves as a comprehensive metric for assessing both detection accuracy and robustness. On the test set, the model achieved a Precision of 0.876, a Recall of 0.868, an mAP@50 of 0.893, and an mAP@50–95 of 0.761. These results indicate that the model demonstrates strong performance in both localization and classification of disease targets, making it well-suited for high-quality disease detection in real-world agricultural scenarios.

\subsection{AgriDoctor Evaluation}
\subsubsection{Setting}
To comprehensively evaluate the performance of AgriDoctor across its core tasks, we manually constructed a balanced test set based on the AgriMM dataset. This test set contains 300 samples, equally distributed across three tasks: disease classification, disease detection, and knowledge-grounded question answering, with 100 instances for each task. All test samples were excluded from model training to ensure unbiased evaluation. The 1:1:1 task ratio ensures balanced representation and enables reliable assessment of both task-specific and overall performance. 

Inspired by the automatic evaluation methods used in LLaVA and LLaVA-Med~\cite{li2023llava}, we adopted the open-source large language model DeepSeek-V3 as the evaluator. For classification and detection tasks, DeepSeek-V3 evaluates the model output against reference answers along two dimensions: \textit{semantic consistency} and \textit{information completeness}, each scored from 0 to 1, with the task score computed as their average.
To obtain the final performance score, we normalize the scores across all tasks and compute the overall average, which serves as the unified metric of model capability.

To provide high-quality reference answers, we manually constructed gold standards for classification and detection tasks. For the knowledge-grounded QA task, reference answers were generated using GPT-4o-mini, based on each question and its corresponding AgriMM knowledge entry. DeepSeek-V3 then compared the candidate model outputs with these references to assess their alignment in terms of both semantic fidelity and content completeness.
We compared AgriDoctor with several strong baselines, including Qwen2.5-VL-72B-Instruct and GPT-4o-mini, to contextualize its performance in both multimodal and textual agricultural reasoning tasks.

\subsubsection{Evaluation Metrics}

\textbf{Semantic Consistency}(SC): Measures the semantic alignment between the model output and the expert-validated reference answers. It also reflects the accuracy of the model in classifying visual categories.

\textbf{Information Completeness}(IC): Evaluates the extent to which the model output covers the disease categories and diagnostic evidence. 

\begin{figure*}[t]
\setlength{\abovecaptionskip}{0pt}  
\setlength{\belowcaptionskip}{0pt}  
   \begin{center}
\includegraphics[width=1\textwidth]{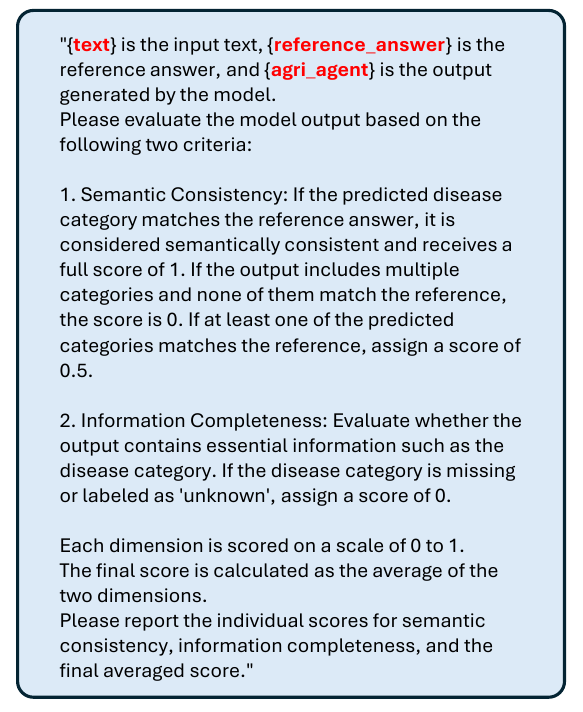}
   \end{center}
   \caption{Classification Evaluation Criteria for Deepseek-V3} 
   \label{fig:ds1}
\end{figure*}

\begin{figure*}[t]
\setlength{\abovecaptionskip}{0pt}  
\setlength{\belowcaptionskip}{0pt}  
   \begin{center}
\includegraphics[width=1\textwidth]{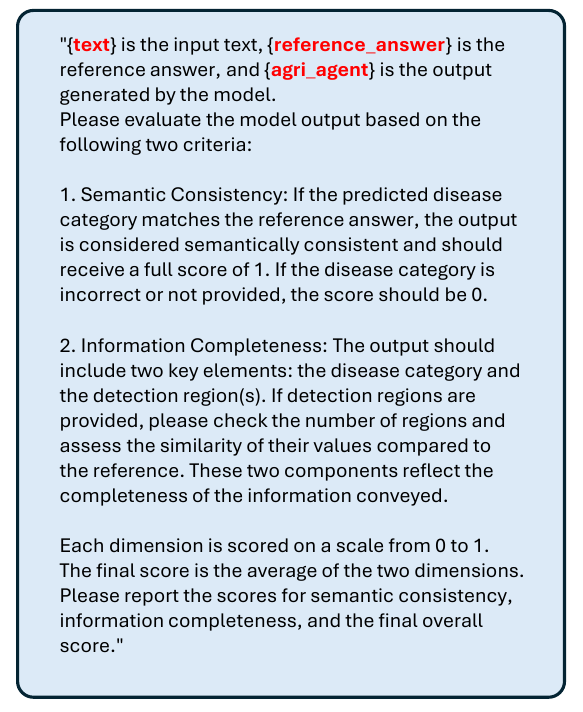}
   \end{center}
   \caption{Object Detection Evaluation Criteria for Deepseek-V3} 
   \label{fig:ds2}
\end{figure*}

\begin{figure*}[t]
\setlength{\abovecaptionskip}{0pt}  
\setlength{\belowcaptionskip}{0pt}  
   \begin{center}
\includegraphics[width=1\textwidth]{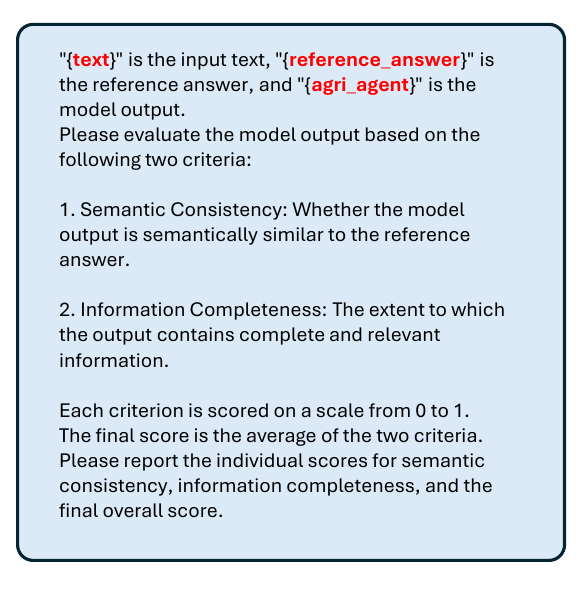}
   \end{center}
   \caption{Expert knowledge on plant diseases Evaluation Criteria for Deepseek-V3} 
   \label{fig:ds3}
\end{figure*}

The evaluation criteria of Deepseek-V3 on the classification task, object detection task, and expert knowledge-based question answering task for crop diseases are illustrated in Figure~\ref{fig:ds1}, Figure~\ref{fig:ds2}, and Figure~\ref{fig:ds3}, respectively. While using DeepSeek-V3 as an automatic evaluator offers a convenient and scalable assessment mechanism, it may introduce potential biases or hallucinations inherent to language models. To mitigate such risks, we additionally performed \textbf{human verification} on a subset of the evaluation results to ensure the reliability and validity of the scoring.

\begin{table}[h]
\centering

\resizebox{\textwidth}{!}{
\begin{tabular}{l|ccc|ccc|ccc|c}
\toprule
\multirow{2}{*}{Model} & \multicolumn{3}{c|}{Classification} & \multicolumn{3}{c|}{Detection} & \multicolumn{3}{c|}{Knowledge Retrieval} & Overall \\
& SC & IC & Avg & SC & IC & Avg & SC & IC & Avg & Score \\
\midrule
Qwen2.5-VL & 0.5200 & 0.9200   & 0.7200  & 0.4360 & 0.4138 & 0.4249 & 0.7640 & 0.9020 & 0.8330 & 0.6593 \\
GPT-4o-mini             & 0.5400 & \textbf{0.9375} & 0.7388  & 0.5665 & 0.4650 & 0.5160 & 0.7760 & 0.9400 & 0.8580 & 0.7043 \\
AgriDoctor              & \textbf{0.8000} & 0.8147 & \textbf{0.8074}  & \textbf{0.8005} & \textbf{0.8093} & \textbf{0.8050} & \textbf{0.9650} & \textbf{0.9880} & \textbf{0.9765} & \textbf{0.8630} \\
\bottomrule
\end{tabular}

}
\caption{Evaluation results on three core agricultural tasks: disease classification, lesion detection, and knowledge-grounded question answering. Each task is evaluated using two metrics: semantic consistency (SC) and information completeness (IC). The final overall score is calculated as the normalized average across all tasks. \textbf{Bold} indicates the best performance in each metric.}
\label{tab:overall-performance}
\end{table}

\subsubsection{Overall Performance}

As shown in Table~\ref{tab:overall-performance}, AgriDoctor consistently achieves the highest scores across all tasks, with an overall performance score of 0.8630, surpassing GPT-4o-mini and Qwen2.5-VL-72B Instruct.
In the classification task, AgriDoctor attains a total score of 0.8074, benefiting from both high semantic consistency and strong information completeness. In the detection task, the model achieves 0.8050, outperforming the baselines in both identifying the correct disease categories and localizing lesion areas. For knowledge retrieval, AgriDoctor obtains a score of 0.9765, demonstrating its effectiveness in understanding complex agricultural queries and retrieving relevant expert-level content from the domain knowledge base.
These results clearly indicate that AgriDoctor delivers strong and consistent performance across all evaluated tasks. Its superiority is particularly evident in knowledge-intensive and multimodal scenarios, validating the effectiveness of its modular architecture, task-specific tool design, and domain-adapted training strategy.

\begin{table}[h]
\centering

\begin{tabular}{l|ccc}
\toprule
Model & SC & IC & Score \\
\midrule
Qwen2.5-14B-Instruct (fine-tuned) & 0.7077 & 0.6760  & 0.6919 \\
Qwen-Max                          & 0.8620 & 0.9770 & 0.9195 \\
AgriDoctor                        & \textbf{0.9650} & \textbf{0.9880} & \textbf{0.9765} \\
\bottomrule
\end{tabular}
\caption{Evaluation results on the agricultural knowledge QA task.}
\label{tab:2}
\end{table}

\subsubsection{Knowledge QA Evaluation}

To further assess the capability of AgriDoctor in agricultural knowledge-based question answering, we constructed a text-only evaluation benchmark aimed at evaluating the model’s reasoning and knowledge integration abilities within the agricultural domain. The comparison involved two representative baselines: a general-purpose large language model, and a domain-adapted model fine-tuned using instruction data.
The fine-tuned model was developed as follows. Based on structured domain knowledge from the AgriMM dataset, we generated 45{,}000 high-quality QA pairs. To ensure fair evaluation and avoid data leakage, any samples overlapping with the test set were removed prior to training. The resulting dataset was then used to perform instruction fine-tuning on the Qwen2.5-14B-Instruct model.

As shown in Table~\ref{tab:2}, compared to the general-purpose model Qwen-Max , AgriDoctor demonstrates superior understanding and coverage of domain-specific content. Additionally, AgriDoctor significantly outperforms the fine-tuned Qwen2.5-14B-Instruct model , highlighting the effectiveness of its modular knowledge retrieval and response generation pipeline.
These results confirm that AgriDoctor not only captures domain-relevant semantics more accurately but also delivers more complete and informative answers. This validates its design as a reliable and knowledge-grounded system for agricultural question answering.
\section{Conclusion}
In this work, we present AgriDoctor, a modular and extensible multimodal framework tailored for intelligent crop disease diagnosis and agricultural knowledge interaction. To support fine-grained training and evaluation, we construct AgriMM. Extensive experiments demonstrate that AgriDoctor, when trained on AgriMM, significantly outperforms existing state-of-the-art vision-language models across multiple agricultural tasks. We believe that AgriDoctor and AgriMM together will serve as a strong foundation for future research in domain-specific vision-language modeling.
\section{Limitations}
Despite the strong performance of AgriDoctor across a range of agricultural tasks, several limitations remain. First, although AgriMM provides a large-scale and diverse set of annotated data, it may still lack sufficient coverage of rare crop species, uncommon disease variants, and region-specific agricultural conditions, which could limit the model's generalizability in underrepresented scenarios. Second, while the current framework supports both Chinese and English, its multilingual capability is still limited, thereby constraining its applicability in low-resource language regions. Addressing these limitations represents a promising direction for future research, with the goal of developing a more comprehensive, adaptive, and globally deployable agricultural AI assistant.
\clearpage
{
\small
\bibliography{mypaper}
\bibliographystyle{unsrt}
}
\clearpage
\section{Appendix}

\subsection{ Disease and Pest Expertise of Wheat Leaf Rust}
\textbf{Introduction}
Wheat leaf rust, also known as yellow rust, is a disease caused by Puccinia recondita f. sp. tritici, affecting wheat. This disease occurs in many wheat-growing regions worldwide, including North America, Europe, Asia, Australia, and Africa. In China, it is particularly prevalent in the southwest and the Yangtze River basin, with significant occurrences also reported in parts of North and Northeast China.

\textbf{Symptoms}
Wheat leaf rust primarily affects wheat leaves, producing pustule-like lesions, and rarely affects the leaf sheath or stem. In the early stages, round or near-round orange-red pustules (uredinia) appear on the leaves. When these break open, they release yellowish-brown powder, which are the urediniospores. In later stages, dark brown to deep brown, elliptical telia form on the leaf undersides, which do not rupture upon maturity.

\textbf{Transmission}
The pathogen mainly overwinters and oversummers as urediniospores or mycelia on wheat and gramineous weeds. Oversummering fungi can cause disease in autumn seedlings. In spring, the overwintering urediniospores directly infect wheat, or spores may be transported from distant areas via wind, initiating local infections. The pathogen continues to reproduce and re-infect wheat multiple times throughout the season.

\textbf{Impact}
Wheat leaf rust causes early yellow spots on leaves, sheaths, and stems, which merge into larger patches as the disease progresses, forming rust-colored pustules characteristic of urediniospore aggregation. In later stages, black pustules—indicative of teliospore aggregation—may appear. Infected plants suffer severe impacts, including reduced photosynthesis, significant water loss, poor grain filling, early leaf senescence, and ultimately, considerable yield reduction.

\textbf{Control Measures}
\begin{itemize}[leftmargin=*]
\item Agricultural Control: Promote and utilize resistant (or tolerant) wheat varieties. Use a diverse range of cultivars and rotate them to prevent the breakdown of resistance due to new physiological races. Improve cultivation practices, such as thorough tillage, weed removal, and eliminating volunteer wheat to reduce pathogen reservoirs. Avoid early sowing in regions prone to autumn infections to decrease overwintering inoculum. Optimize planting density and fertilization, and avoid excessive or late nitrogen application.

\item Chemical Control: Apply seed treatments before sowing. For example, coat seeds with 0.2\% of a 3\% epoxiconazole suspension or 0.06\% of a 6\% tebuconazole suspension by seed weight. These treatments can also control common bunt, powdery mildew, and other rust diseases.

\item Integrated Control: In China, wheat leaf rust is mainly managed through integrated strategies, prioritizing resistant cultivar deployment, supported by optimized agronomic practices and chemical control. This approach effectively limits the spread of urediniospores and ensures successful disease management.
\end{itemize}
By implementing these integrated strategies, the occurrence and spread of wheat leaf rust can be effectively controlled, ensuring healthy wheat production.
\subsection{Disease Classification Tool}

During training, we adopt a weighted cross-entropy loss to address the long-tailed distribution of class frequencies. The weight $w_y$ assigned to each ground-truth class $y$ is defined as:
\begin{equation}
w_y = \min\left(\frac{N}{N_y}, 10\right),
\end{equation}
where $N$ is the total number of training samples and $N_y$ is the number of samples belonging to class $y$. The overall classification loss is then formulated as:
\begin{equation}
L(p, y) = -w_y \cdot \log\left(\frac{\exp(p_y)}{\sum_{j=1}^{C} \exp(p_j)}\right),
\end{equation}
where $C$ is the total number of classes, $p = [p_1, p_2, \dots, p_C]$ is the logit vector output by the model, and $y \in \{1, 2, \dots, C\}$ is the true class label. This loss formulation penalizes misclassification of minority classes more strongly, while capping the weight to avoid overcompensation.

\subsection{Example}

\begin{figure*}[ht]
\setlength{\abovecaptionskip}{0pt}  
\setlength{\belowcaptionskip}{0pt}  
   \begin{center}
\includegraphics[width=1\textwidth]{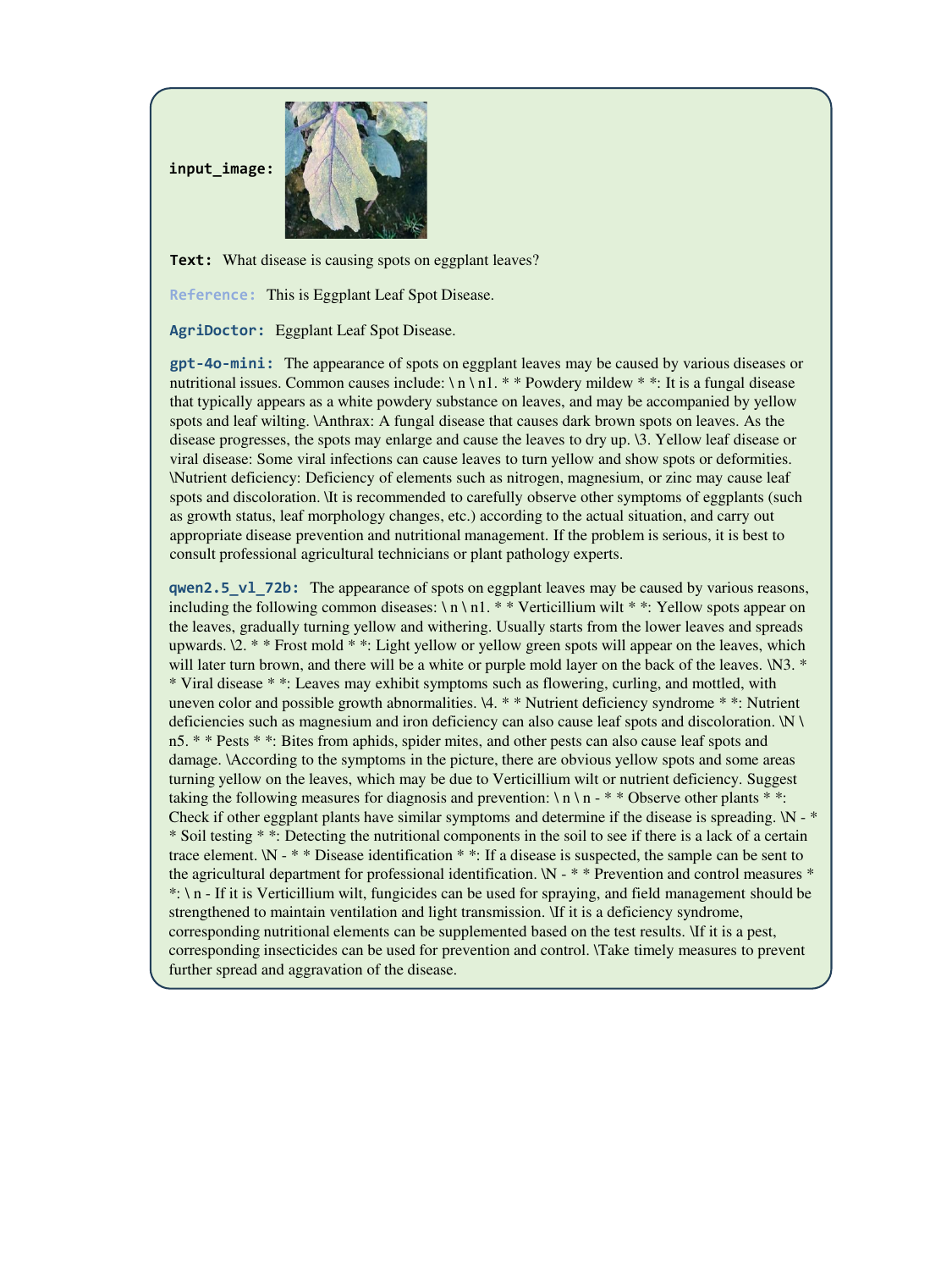}
   \end{center}
   \caption{Comparative Diagnosis Results from AgriDoctor and Baseline Models on Eggplant Leaf Spot Disease } 
   \label{fig:3}
\end{figure*}

\clearpage
\end{document}